\documentclass[letterpaper, 10 pt, conference]{ieeeconf}  

\IEEEoverridecommandlockouts                              

\usepackage{booktabs}
\usepackage{graphicx}
\usepackage{subcaption}
\usepackage[font=small]{caption}
\usepackage{todonotes}
\usepackage{colortbl}
\usepackage{siunitx}
\usepackage{rotating}
\usepackage{multirow}

\usepackage{float}
\newfloat{test}{htbp}{loc}
\usepackage{hyperref}

\usepackage{graphicx}
\usepackage{etoolbox}


\begin{document}

\title{\LARGE \bf
Semantic Masking and Visual Feature Matching for Robust Localization
}

\author{Luisa Mao$^{1, 2}$, Ryan Soussan$^{2}$, Brian Coltin$^{2}$, Trey Smith$^{2}$, Joydeep Biswas$^{1}$
\thanks{*The NASA Game Changing Development Program (Space Technology Mission Directorate) provided funding for this work.}
\thanks{The authors are with  $^{1}$the Department of Computer Science at the University of Texas at Austin and $^{2}$the NASA Ames Research Center 
{\tt\footnotesize \{luisa.mao, joydeepb\}@utexas.edu} \newline
{\tt\footnotesize  \{ryan.soussan, brian.coltin, trey.smith\}@nasa.gov}}
}

\maketitle

\begin{abstract}


We are interested in long-term deployments of autonomous robots to aid astronauts with maintenance and monitoring operations in settings such as the International Space Station. Unfortunately, such environments tend to be highly dynamic and unstructured, and their frequent reconfiguration poses a challenge for robust long-term localization of robots.
 Many state-of-the-art visual feature-based localization algorithms are not robust towards spatial scene changes, and SLAM algorithms, while promising, cannot run within the low-compute budget available to space robots. To address this gap, we present a computationally efficient semantic masking approach for visual feature matching that improves the accuracy and robustness of visual localization systems during long-term deployment in changing environments. Our method introduces a lightweight check that enforces matches to be within long-term static objects and have consistent semantic classes.  We evaluate this approach using both map-based relocalization and relative pose estimation and show that it improves Absolute Trajectory Error (ATE) and correct match ratios on the publicly available Astrobee dataset. While this approach was originally developed for microgravity robotic freeflyers, it can be applied to any visual feature matching pipeline to improve robustness.

\end{abstract}

\section{INTRODUCTION}

\begin{figure}[!t]
    \centering
    \vspace{0.2cm}     \includegraphics[width=0.49\textwidth]{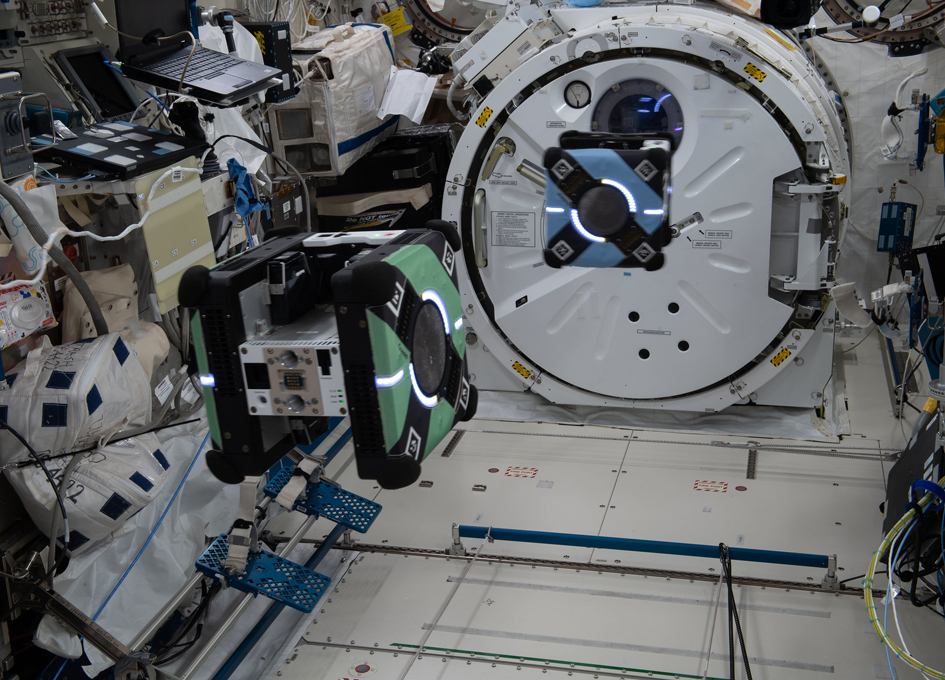} 
    \caption{Astrobee free-flying robots roaming the ISS during an activity. Background objects such as laptops, wires, and cargo bags are often moved between flights and can cause localization errors for the robots.
    }
    \label{fig:astrobees_image}
\end{figure}

Accurate and robust localization is required for reliable long-term robot autonomy. In environments with dynamic or movable objects, place recognition can be challenging as scene consistency is often assumed. The International Space Station (ISS) is an example of such an environment, and the Astrobee robots \cite{smith2016astrobee} operating onboard face constant changes as objects such as cargo bags, wires, laptops, and racks are introduced or rearranged as displayed in Fig.~\ref{fig:astrobees_image}. 
Increasing map matching robustness in the presence of environmental differences would enable more lifelong autonomy for these and other robots.
  \par

Localization for the Astrobee robots is made possible by a specialized system which can handle the microgravity, constricted modules and  planar, repeated scenes of the ISS. As the Astrobee is limited by compute, maps must be pre-built offline. The remote nature of the ISS makes it difficult to remap frequently enough to capture changes, so there are often discrepancies between the map and deployment environment. Additional challenges of the ISS, such as the limited space to move in, planar scenes, and monocular camera images, cause many state-of-the-art visual feature-matching approaches, including ORBSLAM3\cite{campos2021orb}, to fail. The lack gravity and noisy IMU data also preclude other well-known localization systems, such as MAPLAB 2.0 \cite{cramariuc2022maplab} which has ingrained assumptions about gravity. On top of this, these approaches (along with other more recent and robust algorithms) are too computationally intensive to run on the Astrobee, whose compute platform \cite{smith2016astrobee} is roughly 10 times slower than an Intel i9-9980HK 2.4 GHz CPU, and of which only a single core is available for the graph-based localizer.

\begin{figure}[!t]  
    \centering
    \includegraphics[width=0.48\textwidth]{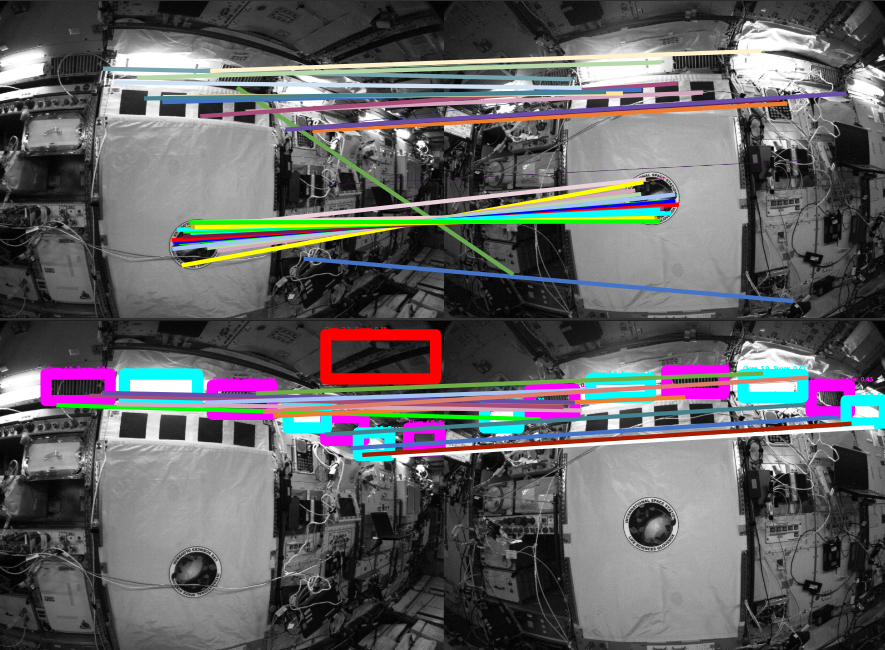}  
    \caption{feature matching with and without bounding boxes. Horizontal image pairs taken several years apart display multiple scene changes, including a rotated ISS flag that causes faulty associations and a failed relative pose estimate in the top image pair. With semantic masks applied to the matches (bottom image pair), detections of stable scene elements including vents (purple), lights (blue), and handrails (red) enable the pruning of faulty associations due to environment changes and successful relative pose estimation.}
    \label{fig:flag_match}
\end{figure}

We are therefore interested in methods which are:
1) Computationally inexpensive, 2) Use visual features and are robust to scene changes, and 3) Can be easily added into an existing visual localization framework for ease of integration.

Bounding box-based semantic segmentation can be run relatively efficiently and  provides object level understanding of a visual scene \cite{iansemantics}.
Semantic segmentation generates object classes that can be used to prune dynamic or unstable objects \cite{yu2018ds} and can improve resiliancy to scene changes by detecting stable, static classes and removing those likely to change over time. 


\par
To take advantage of the accuracy of feature-based matches and robustness of using semantics, we present a meta-algorithm that enhances visual feature matching for mapping and localization.
Our contributions include:
\begin{itemize}
\item A semantic masking stage applied to visual feature matching that enforces class consistency between matches using efficient bounding box detections. This approach can be used with any visual SLAM or localization algorithm to improve robustness to scene changes.
\item An evaluation using the publicly available Astrobee ISS dataset \cite{astrodataset} demonstrating increased accuracy and robustness for both map-based pose estimates and relative correspondences in image pairs.  
\end{itemize}
 

\section{RELATED WORK}

\subsection{Geometric Approaches}


ORB-SLAM3 relies on ORB features \cite{rublee2011orb} and a distributed bag of words (DBoW) \cite{galvez2012bags} for place recognition and loop closures. It quantizes the feature space by clustering descriptors into visual words, and queries are made by finding map frames described by similar visual words. 
MabLab collects BRISK \cite{leutenegger2011brisk} or FREAK \cite{alahi2012freak} features to build a sparse map alongside performing online VIO, which is later optimized offline. 
COLMAP \cite{schonberger2016structure} matches SIFT features and performs bundle-adjustment using the matches. While each of these approaches are quite successful at matching images from individual activities or without large changes over time, they ignore semantics of the environment and are prone to matching errors if the surroundings change.  


\subsection{Semantic Approaches}
\subsubsection{Localization}
Miller \emph{et al.} \cite{iansemantics} explore the use of semantic maps for localization, introducing a new mapping technique which constructs 3d heatmaps of object locations from the use of a bounding box object detector in the image space. 
Though adding semantic localization improved accuracy when no map-based visual features were otherwise available, it decreased it when both were accessible. \par
X-View \cite{gawel2018x} uses pixel level semantics to construct descriptors from segmented frames, but does not incorporate geometric features, using only an odometry source for relative pose estimation in addition to the semantic matches.
Similarly, Liu \emph{et al.} \cite{liu2019global} also relies on random walk descriptors to match semantic objects. Both of these approaches rely on dense pixel-level detections that require increased computation and expensive datasets for training. 
\subsubsection{Odometry}
VSO \cite{lianos2018vso} uses dense pixel-level semantics and introduces a semantic likelihood function to optimize semantic reprojection errors for visual odometry. Semantic-Direct Visual Odometry \cite{bao2022semantic} also uses pixel-level semantics, but performs dense alignment of semantic images. An \emph{et al.} \cite{an2017semantic} perform visual odometry using dense semantics to assign weights for sparse reprojection errors based on their semantic classes and similarly to prioritize sampling certain matches during a RANSAC-based essential matrix calculation. They additionally performed semi-dense matching between images using patches matching defined static semantic classes.
\subsubsection{SLAM} 
Wang \emph{et al.} \cite{8708875} demonstrate that semantics could enhance SLAM by integrating YOLO with ORB-SLAM2, evaluating on the Freiburg dataset and with an RGB-D quadcopter system.
We present a different method of integration which requires structural differences to the sparse map and a more in-depth evaluation including ML baselines on data specific to our application, on which out-of-the-box SLAM approaches fail.

Bowman \emph{et al.} \cite{bowman2017probabilistic} integrate image semantics with geometric features in the same SLAM algorithm but decouple these as inputs, relying on map projections into detected semantic bounding boxes and geometric feature tracking between keyframes. Civera \emph{et al.} \cite{civera2011towards} use a map of objects with extracted SURF \cite{bay2008speeded} features, but do not use a semantic detector to filter or classify matches. Instead they rely on a RANSAC projection algorithm to identify any detected objects in new images. 
Kimera \cite{rosinol2020kimera} \cite{chang2021kimera} generates a 3D metric-semantic mesh using image-space detections, but only adds semantics after performing SLAM.


\subsection{Learning Based Matching}

Research into attention-based GNN matchers have produced algorithms such as Superglue \cite{sarlin2020superglue} which reason about the geometry of the scene. However, the ability of Superglue to capture spatial relationships does not help when the spatial relationships between the components of the scene change. In a dynamic environment, not all parts of the scene are useful and a controlled way to select the useful portions is needed.
Additionally, GNNs are difficult to interpret, whereas our approach gives the domain expert control in picking portions of the scene which have semantic meaning.
\par

Erlich \emph{et al.} explore the use of object-level features \cite{10161393} for object matching across large viewpoint changes. 
They find that keypoint-based descriptors used with SuperGlue perform better on images with smaller viewpoint changes, but are not as robust as object-level descriptors when there are large viewpoint changes. 
Whereas Erlich \emph{et al.} focus on robustness towards viewpoint changes for the same scene, we focus on robustness towards changes to the scene itself. Rather than combining objects and keypoints through a match score, we compare approaches using object detection either as preprocessing or post-match filtering, and provide an evaluation of a real pipeline.

\section{METHOD}
     
\begin{figure}
    \centering
    \begin{minipage}{\textwidth}
        \begin{subfigure}{0.24\textwidth}
            \includegraphics[width=\textwidth, height=0.6\textwidth]{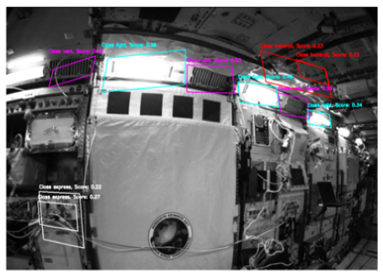}
            \caption{Bounding Boxes}
            \label{fig:bounding_boxes}
        \end{subfigure}%
        \begin{subfigure}{0.24\textwidth}
            \includegraphics[width=\textwidth, height=0.7\textwidth]{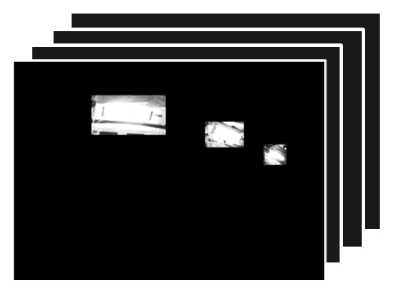}
            \caption{Masked Images}
            \label{fig:masked_images}
        \end{subfigure}
    \end{minipage}
    \begin{subfigure}{0.8\textwidth}
        \includegraphics[width=0.6\textwidth]{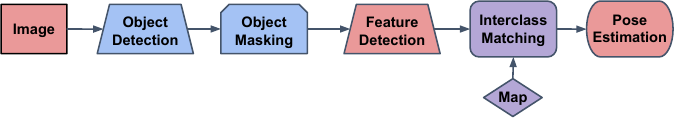}
        \caption{Semantic filtering pipline}
        \label{fig:subfig3}
    \end{subfigure}
    \caption{The semantic image matching pipeline adds semantic segmentation stages in blue to a visual feature matching pipeline in red to improve pose estimation accuracy. The pipeline detects semantic objects in each image (Fig.~\ref{fig:bounding_boxes}) and generates masked image-space regions for each detection in each object class (Fig.~\ref{fig:masked_images}). It then detects visual features in the masked regions and performs matching between features of the same class for each pair of images. Finally, the pipeline estimates the relative pose between the images using the resulting matches.}
    \label{fig:pipeline}
\end{figure}
The semantic image matching pipeline depicted in Fig.~\ref{fig:pipeline} improves upon traditional feature matching approaches by adding semantic filtering on a per class basis to visual feature matches.



\subsection{Object Detection} The semantic object detection stage in the pipeline uses a bounding box object detector fine-tuned on ISS data and with eight defined object classes \cite{iansemantics}. Semantic bounding boxes are displayed in Fig.~\ref{fig:bounding_boxes}, where three classes (vents, lights, and handrails) are detected.

\subsection{Object Masking}
The pipeline generates masks for each image using the detected semantic bounding boxes as shown in Fig.~\ref{fig:masked_images}. Regions without semantic detections are not used for later stages of the matching pipeline. Masking is performed before feature detection to improve runtime as features only need to be calculated in masked regions.

\subsection{Feature Detection}
The matching pipeline uses SURF \cite{bay2008speeded} features and hyperparameters tuned for the ISS for feature detection. The SURF detector relies on a dynamic Hessian threshold which adjusts itself until there are between 1000 and 5000 features extracted for each image \cite{coltin2016localization}. Fig.~\ref{fig:flag_match} shows example detections for an ISS image.

\subsection{Interclass Matching}
\subsubsection{Map}
The interclass matching stage relies on a prebuilt feature map that consists of extracted SURF keypoints and their triangulated 3d positions \cite{coltin2016localization} augmented with semantic labels from the semantic object detector. Only keypoints with valid semantic object detections are retained in the map which drastically reduces the memory usage.


\subsubsection{Feature Matching}
Candidate matching images in the map are obtained for each image using a DBoW query \cite{galvez2012bags}.
The pipeline matches features with the corresponding semantic labels using the FLANN matcher \cite{muja2009fast} with a goodness ratio of 0.7.  
Fig.~\ref{fig:flag_match} depict the image matching results with and without semantic filtering.

\subsection{Pose Estimation}
The pose estimation stage of the matching pipeline uses the perspective-three-point algorithm \cite{gao2003complete} to estimate the camera pose from the 2d-3d matches between the image and map. A RANSAC selection procedure \cite{coltin2016localization} iteratively computes poses using four randomly sampled matches at a time and returns the pose with the most inlier matches.

 

\subsection{Implementation}





The Astroloc relocalization module performs visual-feature matching to a pre-built sparse map to recover pose.
 Due to the importance of this module of the localization pipeline as the only method of recovery should the Astrobee become lost, we choose to integrate the semantic filter into this module.

We evaluate offline, though all of the individual components of the pipeline, including the object detector model, have previously been successfully run on the Astrobee robots.

\section{EXPERIMENTS}
Our experiments show the effects of the semantic filter on 2d-2d matching for both classical and learning-based systems, answering the following questions: \\
\textbf{1) Does the use of semantics improve visual feature matching as used for visual localization?} \\
\textbf{2) What are the effects of the semantic filter on learning-based matching approaches which already incorporate spatial relations?} Though Astrobee is not currently capable of using these techniques, future missions may use more advanced localization techniques that may benefit from semantic masking.
\par To answer these, we evaluate our approach using the Astroloc \cite{soussan2022astroloc} map-based relocalizer with and without semantic filtering. Additionally, we compare the performance of the learning-based image feature matcher Superglue on learned Superpoint \cite{detone2018superpoint} features both with and without semantics.

\subsection{Dataset}

The algorithms are evaluated using eight publicly available datasets from Astrobee deployment in the Japanese Experiment Module (JEM) on the ISS \cite{astrodataset}. Table~\ref{table:sequence_key} shows the key to the sequence names. Our data spans from 2019 to 2022, and covers a variety of activities, viewpoints, and lighting. The repeated deployments in the same contained environment gives an opportunity to observe changes to the scene through time.

\begin{table}[h]
    \centering
        \begin{tabular}{ll |ll| ll}
            \toprule
            1 & tb\_roll  & 4& ff\_return\_journey\_forward  & 6& iva\_kibo\_trans  \\
            2 & tb\_pitch & 5& ff\_return\_journey\_left & 7& iva\_kibo\_rot \\
            3& tb\_yaw & && 8& iva\_kibo\_tag \\
            \bottomrule

 \end{tabular}
    \caption{Key of Number to Sequence Name in Astrobee Dataset}
    \label{table:sequence_key}
\end{table}

\subsection{Visual Localization}
\subsubsection{Metrics}
The Absolute Position Error (APE) in meters and the Absolute Rotation Error (ARE) in degrees are calculated for each relocalized pose in the trajectory. We report the max and median errors along with RMSE, since even a single large failure in relocalization can impact subsequent state estimations. We also calculate the Success Rate (SR), or percentage of localized poses within 0.3m and 5 degrees of groundtruth. These results can be directly compared to the evaluations of SLAM baselines in \cite{astrodataset}.

\subsection{Image Feature Matching}
\subsubsection{Metrics}
Unlike the localization setup where a global pose is recovered from the collected set of 2d-3d matches between a query image and a list of map images, the image feature matching evaluation uses Superglue to find the relative camera pose between pairs of images. Within each trajectory, each image is paired with the single most similar image from the image database and Superglue matches are used to estimate the essential matrix, from which the relative camera pose is recovered. The rotation error in degrees and the translation heading error (the angular difference between the norm of translation vectors) between the estimated and groundtruth extrinsic transformations are reported. We also report the average proportion of correctly matched keypoints (defined by having an epipolar error less than 5e-4) over the entire trajectory.

\subsubsection{Segmentation as Pre-Processing}
Each image pair is segmented and masked, and the masked pairs of images are matched in eight passes according to object class. All matches are collected and the essential matrix is estimated using a five-point relative pose method \cite{nister2004efficient}.
\subsubsection{Segmentation as Post-Processing}
Each image pair is matched with SuperGlue and matches where both keypoints are within bounding boxes of the same semantic class are kept. The filtered matches are again used to estimate the essential matrix.
   
\section{RESULTS}

\subsection{Visual Localization}
\sisetup{round-mode=places,round-precision=4}
\begin{table}[h]
    \setlength{\tabcolsep}{4pt}
    \centering 
     \vspace{0.2cm}
\begin{tabular}{l|rr|rr|rr}
    \toprule
                             & max                  &                      & median               &                      & RMSE                 &                      \\
Seq                     & Baseline  & Semantic  & Baseline  & Semantic  & Baseline  & Semantic  \\
\toprule
1                     & \cellcolor[gray]{.9}\num{0.015403}       & \num{0.015633}       & \num{0.005671}       & \cellcolor[gray]{.9}\num{0.005492}       & \cellcolor[gray]{.9}\num{0.007647}       & \num{0.00846}        \\
2                    & \num{0.026673}       & \cellcolor[gray]{.8}\num{0.013847}       & \num{0.007639}       & \cellcolor[gray]{.8}\num{0.006259}       & \num{0.00953}        & \cellcolor[gray]{.8}\num{0.00735}        \\
3                      & \num{1.350333}       & \cellcolor[gray]{.6}\num{0.376626}       & \cellcolor[gray]{.9}\num{0.032541}       & \num{0.032697}       & \num{0.30133}        & \cellcolor[gray]{.6}\num{0.057787}       \\
\midrule
4 & \num{1.437444}       & \cellcolor[gray]{.8}\num{1.326094}       & \num{1.022843}       & \cellcolor[gray]{.8}\num{1.021239}       & \num{0.978852}       & \cellcolor[gray]{.8}\num{0.970929}       \\
5   & \cellcolor[gray]{.8}\num{1.043529}       & \num{1.066863}       & \num{0.334358}       & \cellcolor[gray]{.8}\num{0.329946}       & \num{0.337499}       & \cellcolor[gray]{.8}\num{0.332322}       \\
\midrule
6             & \num{3.405037}       & \cellcolor[gray]{.6}\num{2.068955}       & \cellcolor[gray]{.9}\num{0.015037}       & \num{0.01563}        & \num{0.42808}        & \cellcolor[gray]{.7}\num{0.259888}       \\
7               &\cellcolor[gray]{.8} \num{1.005764}       & \num{2.277931}       & \num{0.017173}       & \cellcolor[gray]{.8}\num{0.014688}       & \cellcolor[gray]{.8}\num{0.098748}       & \num{0.173117}       \\
8            & \num{1.167855}       & \cellcolor[gray]{.6}\num{0.530592}       & \cellcolor[gray]{.8}\num{0.093657}       & \num{0.102779}       & \num{0.229465}       & \cellcolor[gray]{.8}\num{0.129597}    \\
\bottomrule
\end{tabular}
     \caption{Non-Semantic vs. Semantic relocalization ATE (m) on Astrobee ISS Datasets}
    \label{tab:ate_table}
\end{table}
\begin{table}[h]
    \setlength{\tabcolsep}{4pt}
    \centering
\begin{tabular}{l|rr|rr|rr}
 \toprule
                             & max                  &                      & median               &                      & RMSE                 &                      \\
Seq                     & Baseline  & Semantic  & Baseline  & Semantic  & Baseline  & Semantic  \\
\toprule
1                    & \cellcolor[gray]{.9}\num{0.001764}       & \num{0.002231}       & \cellcolor[gray]{.9}\num{0.00082}        & \num{0.000865}       & \cellcolor[gray]{.9}\num{0.000883}       & \num{0.000961}       \\
2                    & \num{0.006773}       & \cellcolor[gray]{.9}\num{0.006228}       & \cellcolor[gray]{.9}\num{0.002643}       & \num{0.003175}       & \cellcolor[gray]{.9}\num{0.003237}       & \num{0.003351}       \\
3                     & \num{3.141498}       & \cellcolor[gray]{.8}\num{3.069498}       & \num{0.029714}       & \cellcolor[gray]{.8}\num{0.029167}       & \num{1.474307}       & \cellcolor[gray]{.6}\num{0.202956}       \\
\midrule
4 & \cellcolor[gray]{.8}\num{0.184153}       & \num{0.441073}       & \cellcolor[gray]{.9}\num{0.006386}       & \num{0.006457}       & \cellcolor[gray]{.9}\num{0.014822}       & \num{0.021205}       \\
5    & \cellcolor[gray]{.8}\num{0.190478}       & \num{0.204514}       & \num{0.029229}       & \cellcolor[gray]{.9}\num{0.027218}       & \num{0.036205}       & \cellcolor[gray]{.9}\num{0.034287}       \\
\midrule
6             & \num{0.563893}       & \cellcolor[gray]{.8}\num{0.366709}       & \cellcolor[gray]{.9}\num{0.004334}       & \num{0.004954}       & \num{0.076822}       & \cellcolor[gray]{.8}\num{0.046994}       \\
7              &\cellcolor[gray]{.8} \num{0.357659}       & \num{0.637306}       & \num{0.008638}       & \cellcolor[gray]{.9}\num{0.00498}        & \cellcolor[gray]{.9}\num{0.034576}       & \num{0.055242}       \\
8              & \num{0.889333}       & \cellcolor[gray]{.8}\num{0.361835}       & \cellcolor[gray]{.9}\num{0.072258}       & \num{0.085769}       & \num{0.179128}       & \cellcolor[gray]{.8}\num{0.09838}    \\    
\bottomrule
\end{tabular}
     \caption{Non-semantic relocalization v.s. Semantic ARE (deg.) on Astrobee ISS Datasets}
    \label{tab: are_table}
\end{table}
\begin{table}[h!]
    \centering 
\begin{tabular}{l|rr}
 \toprule

Sequence                     & Baseline & Semantic \\
\toprule
1                     & \num{1}              & \num{1}              \\
2                   & \num{1}              & \num{1}              \\
3                      & \num{0.7755}         & \cellcolor[gray]{.8}\num{0.9429}         \\
\midrule
4 & \num{0.0433}         & \num{0.0433}         \\
5    & \num{0.3012}         & \cellcolor[gray]{.8}\num{0.3034}         \\
\midrule
6             & \num{0.4813}         & \cellcolor[gray]{.8}\num{0.544}          \\
7               & \cellcolor[gray]{.8}\num{0.7859}         & \num{0.7103}         \\
8               & \num{0.4803}         & \cellcolor[gray]{.8}\num{0.5263}  \\
\bottomrule
\end{tabular}
 \caption{Relocalization success rates with and without semantics on Astrobee Datasets.}
\label{tab: success_rate_table}
\end{table}
Table \ref{tab:ate_table} displays a reduction in ATE when using semantics for all but two datasets, whereas Table \ref{tab: are_table} shows both approaches attained low ARE. 
To further illustrate performance, we show the success rates for the datasets in Table \ref{tab: success_rate_table}, where semantics improve relocalization for all but one dataset. 
The difference between the Astroloc relocalizer with and without the semantic filter is best observed when there are modular changes to the environment.

\begin{figure}[t!]
\captionsetup[subfigure]{aboveskip=3pt}
    \begin{subfigure}[b]{0.9\linewidth}
        \centering
        \includegraphics[width=.99\linewidth]{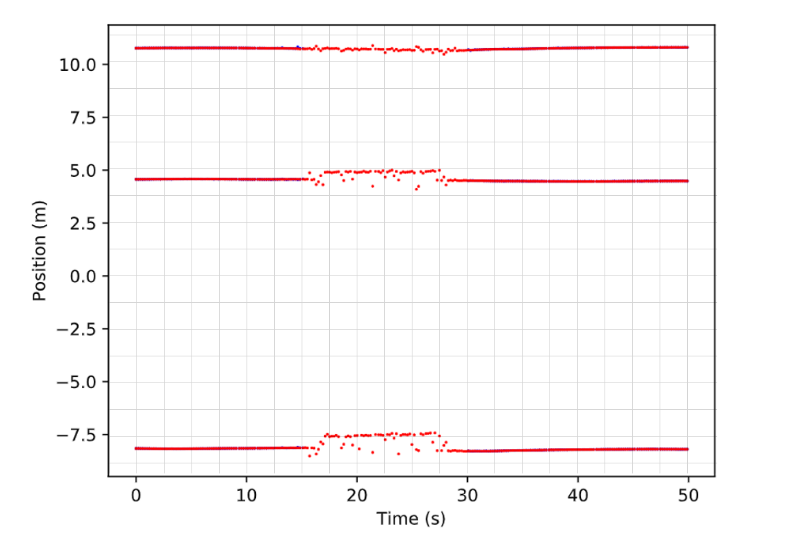} 
        \caption{Baseline relocalizer}
        \label{fig:pose}
    \end{subfigure}
    \begin{subfigure}[b]{0.9\linewidth}
        \centering
        \includegraphics[width=0.99\linewidth]{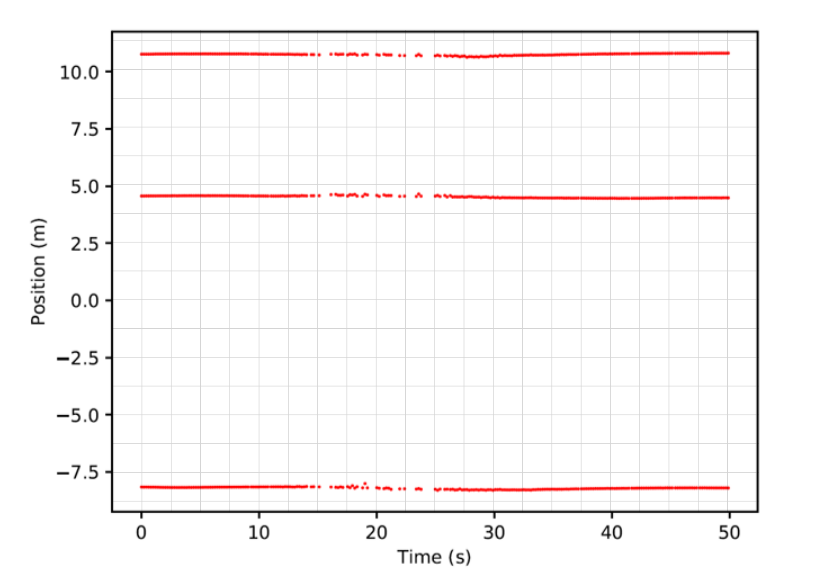} 
        \caption{Semantic relocalizer}
        \label{fig:semantic_pose}
    \end{subfigure}
    \caption{The XYZ position of the Astrobee through time in the tb\_yaw sequence is plotted above. The non-semantic localizer accrues a position offset in the middle of the plot (visible as a discontinuous step) whereas the semantic localizer maintains its fixed position.}
    \label{fig:comparison_plot_position}
\end{figure}

\begin{figure}[t!]
\captionsetup[subfigure]{aboveskip=3pt}
    \begin{subfigure}{0.9\linewidth}
        \centering
        \includegraphics[width=0.99\linewidth]{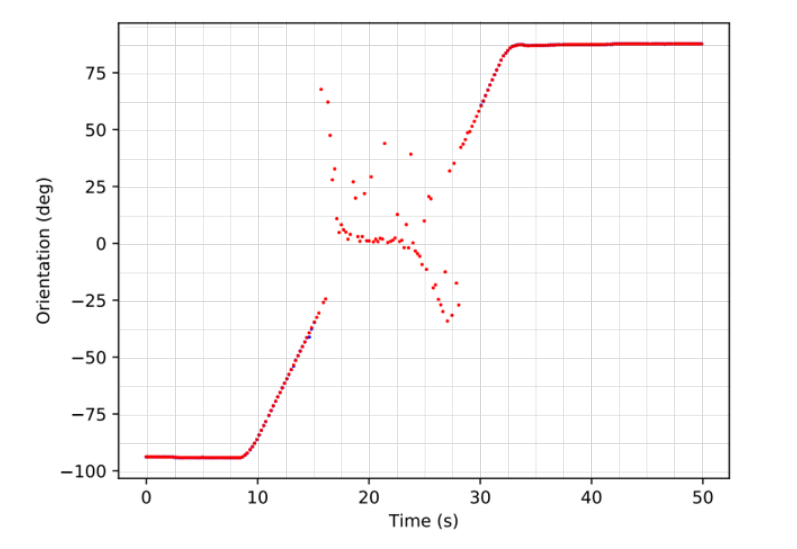}
        \caption{Baseline relocalizer}
        \label{fig:yaw}
    \end{subfigure}
        \begin{subfigure}{0.9\linewidth}
        \centering
        \includegraphics[width=.99\linewidth]{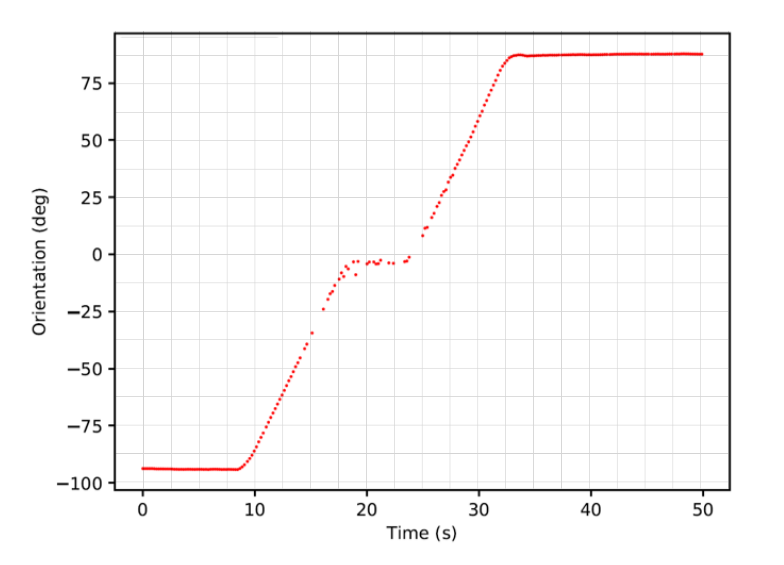}
        \caption{Semantic relocalizer}
        \label{fig:semantic_yaw}
    \end{subfigure}

        \caption{The addition of semantics helps the robot track its orientation during an in-place rotation in the tb\_yaw sequence. Here, the non-semantic relocalizer localizes upside-down and observes a reversed yaw around 15 seconds while the semantic version properly tracks the rotation.}
        \label{fig:comparison_plot_yaw}
\end{figure}

A particularly interesting example occurs in the tb\_yaw sequence, in which the robot spins around its z-axis from facing one end of the JEM to facing the other end. In the middle of its trajectory, the robot observes a flag which has been flipped upside-down and is inconsistent with its prior map as shown in Fig.~\ref{fig:flag_match}. This causes the localized poses from the entire middle portion of its trajectory to be upside down with respect to the map. This is fixed when using semantics as displayed in Fig.~\ref{fig:flag_match}. We further highlight this in Fig.~\ref{fig:comparison_plot_position} and \ref{fig:comparison_plot_yaw} where a position offset in the non-semantic relocalizer and reversed yaw between approximately 15 and 30 seconds are both avoided when using semantics. 

The errors of the iva\_kibo\_trans and iva\_ARtag sequences from years 2022 and 2021 are also lower with the addition of the semantic filter. As shown here, there can be serious failures if changes in the environment are unnoticed and unreflected in the map. For microgravity free-flyers in particular, these issues are exacerbated by the lack of a gravity vector to verify against and the difficulty of creating maps frequently enough to capture changes due to the inaccessibility of the ISS. 

In other datasets, there is negligible difference of the Astroloc relocalization module with ground truth, as the environment and the map are similar enough for the relocalizer to find the robot’s pose. Changes are usually contained within certain portions of the environment, resulting in segments of the trajectory being mislocalized, which is not well-conveyed when the absolute position or rotation error is averaged over the entire trajectory.  In iva\_kibo\_rot, the localization with the semantic filter has greater error than without, since only using features within boxes results in less inliers with which to refine the camera pose.

Though not explicity shown, the two visual localization baselines ORB-SLAM3 and maplab 2.0 were also employed on Astrobee data. Unlike the evaluations in the Astrobee ISS dataset \cite{astrodataset}, ORB-SLAM3 and maplab 2.0 were run in localization mode on a map built several years apart from the evaluation datasets. Both algorithms failed to find loop closures with the previous map and relied only on odometry. Furthermore, maplab 2.0 could not be used without additional engineering effort due to their assumptions about the existence of a gravity vector.

We also note that since we use a map built years apart from the bags used to evaluate, there is a registration difference which causes the success rate of  ff\_return\_journey\_forward to be low, even with the map origin alignment.

\subsection{Image Feature Matching}
\begin{figure}[h]
    \centering
    \begin{subfigure}[b]{0.49\textwidth}
        \includegraphics[width=\textwidth]{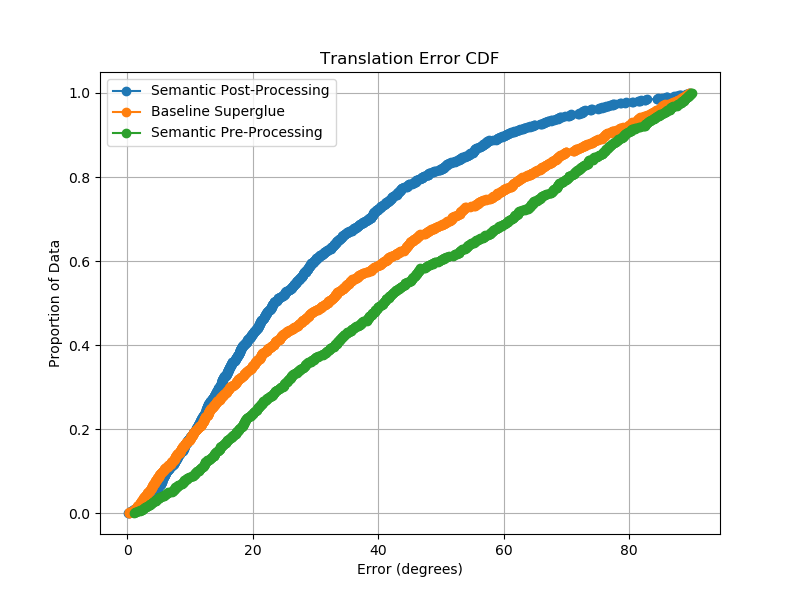}
        \caption{Translation Heading Error (deg)}
        \label{fig:total_cdf_t}
    \end{subfigure}
    \begin{subfigure}[b]{0.49\textwidth}
        \includegraphics[width=\textwidth]{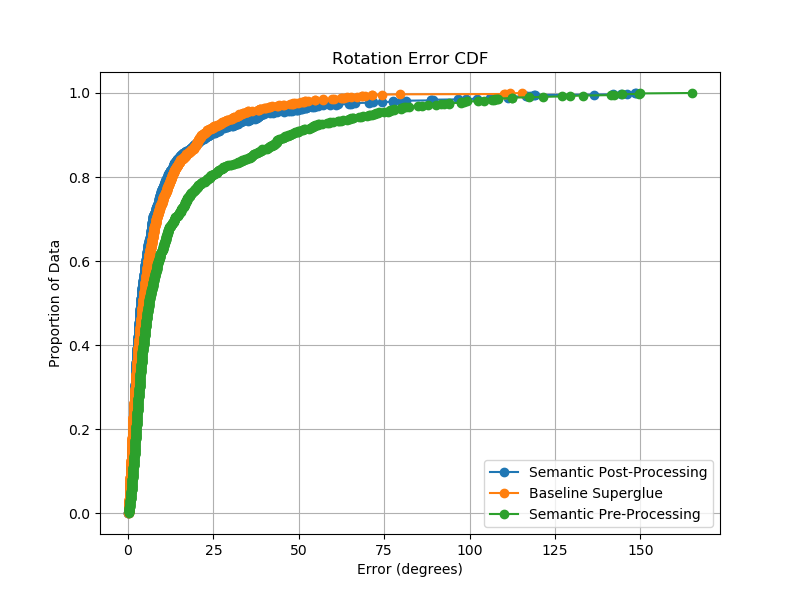}
        \caption{Rotational Error (deg)}
        \label{fig:total_cdf_R}
    \end{subfigure}
    \caption{Translation and Rotation Error CDFs for Superglue with and without semantics. A curve closer to the upper left denotes lower error.}
    \label{fig:total_cdfs}
\end{figure}
\sisetup{round-mode=places,round-precision=4}
\begin{table}[h]
    \setlength{\tabcolsep}{4pt}
     \vspace{0.2cm}
    \centering
        \begin{tabular}{l|rrr|rrr}
            \toprule
            Seq &  \multicolumn{3}{c|}{error t (deg):} & \multicolumn{3}{c}{error R (deg):} \\
            & Baseline & Pre & Post & Baseline & Pre & Post \\
            \toprule
            1 & \num{32.56072} & \num{35.77997} & \cellcolor[gray]{.8}\num{13.92052} & \num{4.140139} & \num{3.53702} & \cellcolor[gray]{.8}\num{2.032448} \\
            2 & \num{22.86025} & \num{39.91108} & \cellcolor[gray]{.8}\num{13.23055} & \num{2.990863} & \num{4.956284} & \cellcolor[gray]{.8}\num{1.955149} \\
            3 & \num{24.5179} & \num{42.2196} & \cellcolor[gray]{.8}\num{14.07822} & \num{5.965214} & \num{10.6487} & \cellcolor[gray]{.8}\num{3.952338} \\
            \midrule
            4 & \num{35.79393} & \num{38.50315} & \cellcolor[gray]{.8}\num{19.02857} & \cellcolor[gray]{.8}\num{3.084002} & \num{6.211718} & \num{3.375606} \\
            5 & \num{35.78241} & \num{45.91112} & \cellcolor[gray]{.8}\num{28.89392} & \cellcolor[gray]{.8}\num{5.95915} & \num{6.912781} & \num{8.312445} \\
            \midrule
            6 & \num{43.83755} & \num{46.10867} & \cellcolor[gray]{.8}\num{33.41859} & \num{21.4706} & \num{27.54666} & \cellcolor[gray]{.8}\num{13.53962} \\
            7 & \cellcolor[gray]{.8}\num{39.78524} & \num{45.77426} & \num{42.56035} & \cellcolor[gray]{.8}\num{14.00608} & \num{22.156} & \num{17.47411} \\
            8 & \cellcolor[gray]{.8}\num{25.41495} & \num{59.04471} & \num{42.09871} & \cellcolor[gray]{.8}\num{12.49089} & \num{67.48147} & \num{21.68465} \\
            \bottomrule
        \end{tabular}
    \caption{Essential matrix estimation errors using Superglue with and without semantic pre and post processing. Since these estimates are not to scale, error is measured in translation and orientation headings.}
    \label{table:super_glue_essential_error}
\end{table}

\begin{table}[!t]
    \centering
    \setlength{\tabcolsep}{5pt}
        \begin{tabular}{l|rrr|rrr}
            \toprule
            Seq &  \multicolumn{3}{c|}{avg correct match ratio:} & \multicolumn{3}{c}{success rate:} \\
            & Baseline & Pre & Post & Baseline & Pre & Post \\
            \toprule
            1 & \num{0.65451} & \cellcolor[gray]{.8}\num{0.812111} & \num{0.688818} & \num{1} & \num{0.435484} & \num{1} \\
            2 & \num{0.726771} & \num{0.790862} & \num{0.745443} & \num{1} & \num{0.98103} & \num{1} \\
            3 & \num{0.520139} & \cellcolor[gray]{.8}\num{0.595147} & \num{0.584439} & \num{1} & \num{0.915} & \num{1} \\
            \midrule
            4 & \num{0.278964} & \cellcolor[gray]{.8}\num{0.37581} & \num{0.323893} & \num{0.748815} & \num{0.63981} & \num{0.748815} \\
            5 & \num{0.305618} & \cellcolor[gray]{.8}\num{0.349949} & \num{0.33982} & \num{0.691304} & \num{0.473913} & \num{0.691304} \\
            \midrule
            6 & \num{0.191862} & \cellcolor[gray]{.8}\num{0.270376} & \num{0.21298} & \num{0.327103} & \num{0.224299} & \num{0.327103} \\
            7 & \num{0.09507} & \cellcolor[gray]{.8}\num{0.104719} & \num{0.098898} & \num{0.161702} & \num{0.12766} & \num{0.161702} \\
            8 & \cellcolor[gray]{.8}\num{0.443206} & \num{0.305245} & \num{0.425435} & \num{0.134868} & \num{0.036184} & \num{0.134868} \\
            \bottomrule
        \end{tabular}
    \caption{Match and success ratios when running Superglue without semantics, using semantics as a pre-processing step, and using semantics as a post-processing step for a sequence of activities.}
    \label{table:super_glue_match_ratios}
\end{table}
The cumulative distribution functions (cdf) in Figures \ref{fig:total_cdf_t} and \ref{fig:total_cdf_R} illustrate the improvements in essential matrix calculation when using semantics with Superglue as a post-processing step. Pre-processing performed worse than baseline, as Superglue always attempts to match 1024 points between images, which can force false matches if different instances of the same object class are detected in each images or heavily concentrate correct matches within boxes. A sample of closely clustered keypoints for essential matrix estimation yields less accurate results than if the keypoints were distributed across the image. 

Table \ref{table:super_glue_match_ratios} displays the improvement in correct match ratios when using semantics as a pre and post processing step compared to not using them. Masking images during pre-processing limits the searchable range for matches, and although the pre-processing results show the largest correct match ratio, this clustering effect yields worse essential matrix estimation compared to post-processed matching as described above. Still, the increase in the ratio of correct matches is still valuable for other applications. If the 3d landmarks have already been triangulated, 2d-2d matching before PnP (as is done in most visual localization pipelines) could be improved with using semantics. Essential matrix estimation performance is displayed in Table \ref{table:super_glue_essential_error}, which again shows semantics as a post-processing step outperforming the baseline Superglue approach and semantics as a pre-processing step. \par
Since the interior of the JEM is shaped as a rectangualar prism, images consist of mostly planar surfaces. Additionally, the microgravity environment results in many query-map image pairs having only relative rotation motion between the camera poses. Despite making high ratios of correct matches on most sequences, these degenerate cases can cause high errors when estimating the essential matrix, especially the translation component. For this reason, Superglue results have much higher errors than the Astroloc evaluation method (where pose can be directly recovered from previously 3d-triangulated points using PnP). 



\section{CONCLUSIONS}
We have presented a lightweight semantic consistency check for visual feature matching that improves the robustness of localization performance. We have shown that enforcing consistent semantic classes for feature matches improves both relocalization performance and essential matrix calculation as evaluated on a dataset of eight Astrobee activities on the ISS. 

As this method is designed to be computationally efficient, we additionally plan to deploy and test our semantic relocalization approach on the Astrobee robots during future ISS activities to improve their resilience to environment changes. 



In future work, we wish to explore using movable object detections as negative matches, and weighting feature matches based on their semantics or lack there of. Additionally, we are interested in further using semantic results to perform informed map updates on an object level. \par




\section*{ACKNOWLEDGMENT}
We would like to thank Ian D. Miller and Suyoung Kang for supporting this work. We would also like to thank Marina Gouviea Moreira for her assistance during testing in the Granite Lab and the rest of the Astrobee Facilities team for their help.

{\footnotesize
\bibliographystyle{IEEEtran}
\bibliography{citations}}

\begin{thebibliography}{10}
\providecommand{\url}[1]{#1}
\csname url@samestyle\endcsname
\providecommand{\newblock}{\relax}
\providecommand{\bibinfo}[2]{#2}
\providecommand{\BIBentrySTDinterwordspacing}{\spaceskip=0pt\relax}
\providecommand{\BIBentryALTinterwordstretchfactor}{4}
\providecommand{\BIBentryALTinterwordspacing}{\spaceskip=\fontdimen2\font plus
\BIBentryALTinterwordstretchfactor\fontdimen3\font minus
  \fontdimen4\font\relax}
\providecommand{\BIBforeignlanguage}[2]{{%
\expandafter\ifx\csname l@#1\endcsname\relax
\typeout{** WARNING: IEEEtran.bst: No hyphenation pattern has been}%
\typeout{** loaded for the language `#1'. Using the pattern for}%
\typeout{** the default language instead.}%
\else
\language=\csname l@#1\endcsname
\fi
#2}}
\providecommand{\BIBdecl}{\relax}
\BIBdecl

\bibitem{smith2016astrobee}
T.~Smith, J.~Barlow, M.~Bualat, T.~Fong, C.~Provencher, H.~Sanchez, E.~Smith
  \emph{et~al.}, ``Astrobee: A new platform for free-flying robotics on the
  {International Space Station},'' in \emph{Int. Symp. on Artificial
  Intelligence, Robotics and Automation in Space}, 2016.

\bibitem{campos2021orb}
C.~Campos, R.~Elvira, J.~J.~G. Rodr{\'\i}guez, J.~M. Montiel, and J.~D.
  Tard{\'o}s, ``Orb-slam3: An accurate open-source library for visual,
  visual--inertial, and multimap slam,'' \emph{IEEE Transactions on Robotics},
  vol.~37, no.~6, pp. 1874--1890, 2021.

\bibitem{cramariuc2022maplab}
A.~Cramariuc, L.~Bernreiter, F.~Tschopp, M.~Fehr, V.~Reijgwart, J.~Nieto,
  R.~Siegwart, and C.~Cadena, ``maplab 2.0--a modular and multi-modal mapping
  framework,'' \emph{IEEE Robotics and Automation Letters}, vol.~8, no.~2, pp.
  520--527, 2022.

\bibitem{iansemantics}
I.~Miller, R.~Soussan, B.~Coltin, T.~Smith, and V.~Kumar, ``Robust semantic
  mapping and localization on a free-flying robot in microgravity,'' in
  \emph{2022 IEEE International Conference on Robotics and Automation
  (ICRA)}.\hskip 1em plus 0.5em minus 0.4em\relax IEEE, 2022.

\bibitem{yu2018ds}
C.~Yu, Z.~Liu, X.-J. Liu, F.~Xie, Y.~Yang, Q.~Wei, and Q.~Fei, ``Ds-slam: A
  semantic visual slam towards dynamic environments,'' in \emph{2018 IEEE/RSJ
  international conference on intelligent robots and systems (IROS)}.\hskip 1em
  plus 0.5em minus 0.4em\relax IEEE, 2018, pp. 1168--1174.

\bibitem{astrodataset}
S.~Kang, R.~Soussan, D.~Lee, B.~Coltin, A.~M. Vargas, M.~Moreira, K.~Hamilton,
  R.~Garcia, R.~Bualat, T.~Smith, J.~Barlow, J.~Benavides, E.~Jeong, and
  P.~Kim, ``Astrobee iss free-flyer datasets for space intra-vehicular robot
  navigation research,'' in \emph{2023 IEEE Robotics and Automation Letters
  (RA-L)}, Under Review.

\bibitem{rublee2011orb}
E.~Rublee, V.~Rabaud, K.~Konolige, and G.~Bradski, ``Orb: An efficient
  alternative to sift or surf,'' in \emph{2011 International conference on
  computer vision}.\hskip 1em plus 0.5em minus 0.4em\relax Ieee, 2011, pp.
  2564--2571.

\bibitem{galvez2012bags}
D.~G{\'a}lvez-L{\'o}pez and J.~D. Tardos, ``Bags of binary words for fast place
  recognition in image sequences,'' \emph{IEEE Transactions on Robotics},
  vol.~28, no.~5, pp. 1188--1197, 2012.

\bibitem{leutenegger2011brisk}
S.~Leutenegger, M.~Chli, and R.~Y. Siegwart, ``Brisk: Binary robust invariant
  scalable keypoints,'' in \emph{2011 International conference on computer
  vision}.\hskip 1em plus 0.5em minus 0.4em\relax Ieee, 2011, pp. 2548--2555.

\bibitem{alahi2012freak}
A.~Alahi, R.~Ortiz, and P.~Vandergheynst, ``Freak: Fast retina keypoint,'' in
  \emph{2012 IEEE conference on computer vision and pattern recognition}.\hskip
  1em plus 0.5em minus 0.4em\relax Ieee, 2012, pp. 510--517.

\bibitem{schonberger2016structure}
J.~L. Schonberger and J.-M. Frahm, ``Structure-from-motion revisited,'' in
  \emph{Proceedings of the IEEE conference on computer vision and pattern
  recognition}, 2016, pp. 4104--4113.

\bibitem{gawel2018x}
A.~Gawel, C.~Del~Don, R.~Siegwart, J.~Nieto, and C.~Cadena, ``X-view:
  Graph-based semantic multi-view localization,'' \emph{IEEE Robotics and
  Automation Letters}, vol.~3, no.~3, pp. 1687--1694, 2018.

\bibitem{liu2019global}
Y.~Liu, Y.~Petillot, D.~Lane, and S.~Wang, ``Global localization with
  object-level semantics and topology,'' in \emph{2019 International Conference
  on Robotics and Automation (ICRA)}.\hskip 1em plus 0.5em minus 0.4em\relax
  IEEE, 2019, pp. 4909--4915.

\bibitem{lianos2018vso}
K.-N. Lianos, J.~L. Schonberger, M.~Pollefeys, and T.~Sattler, ``Vso: Visual
  semantic odometry,'' in \emph{Proceedings of the European conference on
  computer vision (ECCV)}, 2018, pp. 234--250.

\bibitem{bao2022semantic}
Y.~Bao, Z.~Yang, Y.~Pan, and R.~Huan, ``Semantic-direct visual odometry,''
  \emph{IEEE Robotics and Automation Letters}, vol.~7, no.~3, pp. 6718--6725,
  2022.

\bibitem{an2017semantic}
L.~An, X.~Zhang, H.~Gao, and Y.~Liu, ``Semantic segmentation--aided visual
  odometry for urban autonomous driving,'' \emph{International Journal of
  Advanced Robotic Systems}, vol.~14, no.~5, p. 1729881417735667, 2017.

\bibitem{8708875}
Y.~Wang and A.~Zell, ``Improving feature-based visual slam by semantics,'' in
  \emph{2018 IEEE International Conference on Image Processing, Applications
  and Systems (IPAS)}, 2018, pp. 7--12.

\bibitem{bowman2017probabilistic}
S.~L. Bowman, N.~Atanasov, K.~Daniilidis, and G.~J. Pappas, ``Probabilistic
  data association for semantic slam,'' in \emph{2017 IEEE international
  conference on robotics and automation (ICRA)}.\hskip 1em plus 0.5em minus
  0.4em\relax IEEE, 2017, pp. 1722--1729.

\bibitem{civera2011towards}
J.~Civera, D.~G{\'a}lvez-L{\'o}pez, L.~Riazuelo, J.~D. Tard{\'o}s, and J.~M.~M.
  Montiel, ``Towards semantic slam using a monocular camera,'' in \emph{2011
  IEEE/RSJ international conference on intelligent robots and systems}.\hskip
  1em plus 0.5em minus 0.4em\relax IEEE, 2011, pp. 1277--1284.

\bibitem{bay2008speeded}
H.~Bay, A.~Ess, T.~Tuytelaars, and L.~Van~Gool, ``Speeded-up robust features
  (surf),'' \emph{Computer vision and image understanding}, vol. 110, no.~3,
  pp. 346--359, 2008.

\bibitem{rosinol2020kimera}
A.~Rosinol, M.~Abate, Y.~Chang, and L.~Carlone, ``Kimera: an open-source
  library for real-time metric-semantic localization and mapping,'' in
  \emph{2020 IEEE International Conference on Robotics and Automation
  (ICRA)}.\hskip 1em plus 0.5em minus 0.4em\relax IEEE, 2020, pp. 1689--1696.

\bibitem{chang2021kimera}
Y.~Chang, Y.~Tian, J.~P. How, and L.~Carlone, ``Kimera-multi: a system for
  distributed multi-robot metric-semantic simultaneous localization and
  mapping,'' in \emph{2021 IEEE International Conference on Robotics and
  Automation (ICRA)}.\hskip 1em plus 0.5em minus 0.4em\relax IEEE, 2021, pp.
  11\,210--11\,218.

\bibitem{sarlin2020superglue}
P.-E. Sarlin, D.~DeTone, T.~Malisiewicz, and A.~Rabinovich, ``Superglue:
  Learning feature matching with graph neural networks,'' in \emph{Proceedings
  of the IEEE/CVF conference on computer vision and pattern recognition}, 2020,
  pp. 4938--4947.

\bibitem{10161393}
C.~Elich, I.~Armeni, M.~R. Oswald, M.~Pollefeys, and J.~Stueckler,
  ``Learning-based relational object matching across views,'' in \emph{2023
  IEEE International Conference on Robotics and Automation (ICRA)}, 2023, pp.
  5999--6005.

\bibitem{coltin2016localization}
B.~Coltin, J.~Fusco, Z.~Moratto, O.~Alexandrov, and R.~Nakamura, ``Localization
  from visual landmarks on a free-flying robot,'' in \emph{2016 IEEE/RSJ
  International Conference on Intelligent Robots and Systems (IROS)}.\hskip 1em
  plus 0.5em minus 0.4em\relax IEEE, 2016, pp. 4377--4382.

\bibitem{muja2009fast}
M.~Muja and D.~G. Lowe, ``Fast approximate nearest neighbors with automatic
  algorithm configuration.'' \emph{VISAPP (1)}, vol.~2, no. 331-340, p.~2,
  2009.

\bibitem{gao2003complete}
X.-S. Gao, X.-R. Hou, J.~Tang, and H.-F. Cheng, ``Complete solution
  classification for the perspective-three-point problem,'' \emph{IEEE
  transactions on pattern analysis and machine intelligence}, vol.~25, no.~8,
  pp. 930--943, 2003.

\bibitem{soussan2022astroloc}
R.~Soussan, V.~Kumar, B.~Coltin, and T.~Smith, ``Astroloc: An efficient and
  robust localizer for a free-flying robot,'' in \emph{2022 International
  Conference on Robotics and Automation (ICRA)}.\hskip 1em plus 0.5em minus
  0.4em\relax IEEE, 2022, pp. 4106--4112.

\bibitem{detone2018superpoint}
D.~DeTone, T.~Malisiewicz, and A.~Rabinovich, ``Superpoint: Self-supervised
  interest point detection and description,'' in \emph{Proceedings of the IEEE
  conference on computer vision and pattern recognition workshops}, 2018, pp.
  224--236.

\bibitem{nister2004efficient}
D.~Nist{\'e}r, ``An efficient solution to the five-point relative pose
  problem,'' \emph{IEEE transactions on pattern analysis and machine
  intelligence}, vol.~26, no.~6, pp. 756--770, 2004.

\end{thebibliography}
\end{document}